%% file: main.tex
\documentclass[10pt,twocolumn,letterpaper]{article}

\usepackage{cvpr}
\usepackage{times}
\usepackage{epsfig}
\usepackage{graphicx}
\usepackage{amsmath}
\usepackage{amssymb}

\usepackage{times}
\usepackage{epsfig}
\usepackage{graphicx}
\usepackage{amsmath}
\usepackage{amssymb}
\usepackage{multirow}
\usepackage{colortbl}
\usepackage{color}
\usepackage{threeparttable}
\usepackage{booktabs}
\usepackage{adjustbox}
\usepackage{subfig}
\usepackage{caption}

\newcommand{\RR}{\mathbb{R}}
\newcommand{\FF}{\mathbf{F}}

\usepackage[breaklinks=true,bookmarks=false]{hyperref}

\cvprfinalcopy 

\def\cvprPaperID{****} 

\begin{document}

\title{PointFlow: Flowing Semantics Through Points for Aerial Image Segmentation}

\author{
Xiangtai Li$^1$\thanks{The first two authors contribute equally. Email: lxtpku@pku.edu.cn}, 
Hao He$^2$$^,$$^3$$^*$, 
Xia Li$^4$,
Duo Li$^5$, 
Guangliang Cheng$^6$,
\\
Jianping Shi$^6$,
Lubin Weng$^2$,
Yunhai Tong$^1$,
Zhouchen Lin$^1$
\\[0.2cm] 
$ ^1$ Key Laboratory of Machine Perception (MOE), Peking University \\
$ ^2$ NLPR, Institute of Automation, Chinese Academy of Sciences \\
$ ^3$ School of Artificial Intelligence, University of Chinese Academy of Sciences \\
$ ^4$ ETH Zurich
$ ^5$ HUKST 
$ ^6$ SenseTime Research
}

\maketitle

\begin{abstract}
    Aerial Image Segmentation is a particular semantic segmentation problem and has several challenging characteristics that general semantic segmentation does not have. There are two critical issues: The one is an extremely foreground-background imbalanced distribution, and the other is multiple small objects along with the complex background. Such problems make the recent dense affinity context modeling perform poorly even compared with baselines due to over-introduced background context. To handle these problems, we propose a point-wise affinity propagation module based on the Feature Pyramid Network (FPN) framework, named PointFlow. Rather than dense affinity learning, a sparse affinity map is generated upon selected points between the adjacent features, which reduces the noise introduced by the background while keeping efficiency. In particular, we design a dual point matcher to select points from the salient area and object boundaries, respectively. Experimental results on three different aerial segmentation datasets suggest that the proposed method is more effective and efficient than state-of-the-art general semantic segmentation methods. Especially, our methods achieve the best speed and accuracy trade-off on three aerial benchmarks. Further experiments on three general semantic segmentation datasets prove the generality of our method. Code will be provided in (\url{https://github.com/lxtGH/PFSegNets}).
\end{abstract}

\input{1introduction}
\input{2relatedwork}
\input{3method}
\input{4experiment}
\input{5conclusion}
\input{7ack}
\input{6sub}

{\small
\bibliographystyle{ieee}
\bibliography{egbib}
}

\end{document}

%% file: 1introduction.tex
\section{Introduction}
High spatial resolution (HSR) remote sensing images contain various geospatial objects, including airplanes, ships, vehicles, buildings, etc. Understanding these objects from HSR remote sensing imagery has great practical value for urban monitoring and management. Aerial Image segmentation is an important task in remote sensing understanding that can provide semantic and localization information cues for interest targets. It is a specific semantic segmentation task that aims to assign a semantic category to each image pixel. 

\begin{figure}[t]
	\centering
\includegraphics[width=1.0\linewidth]{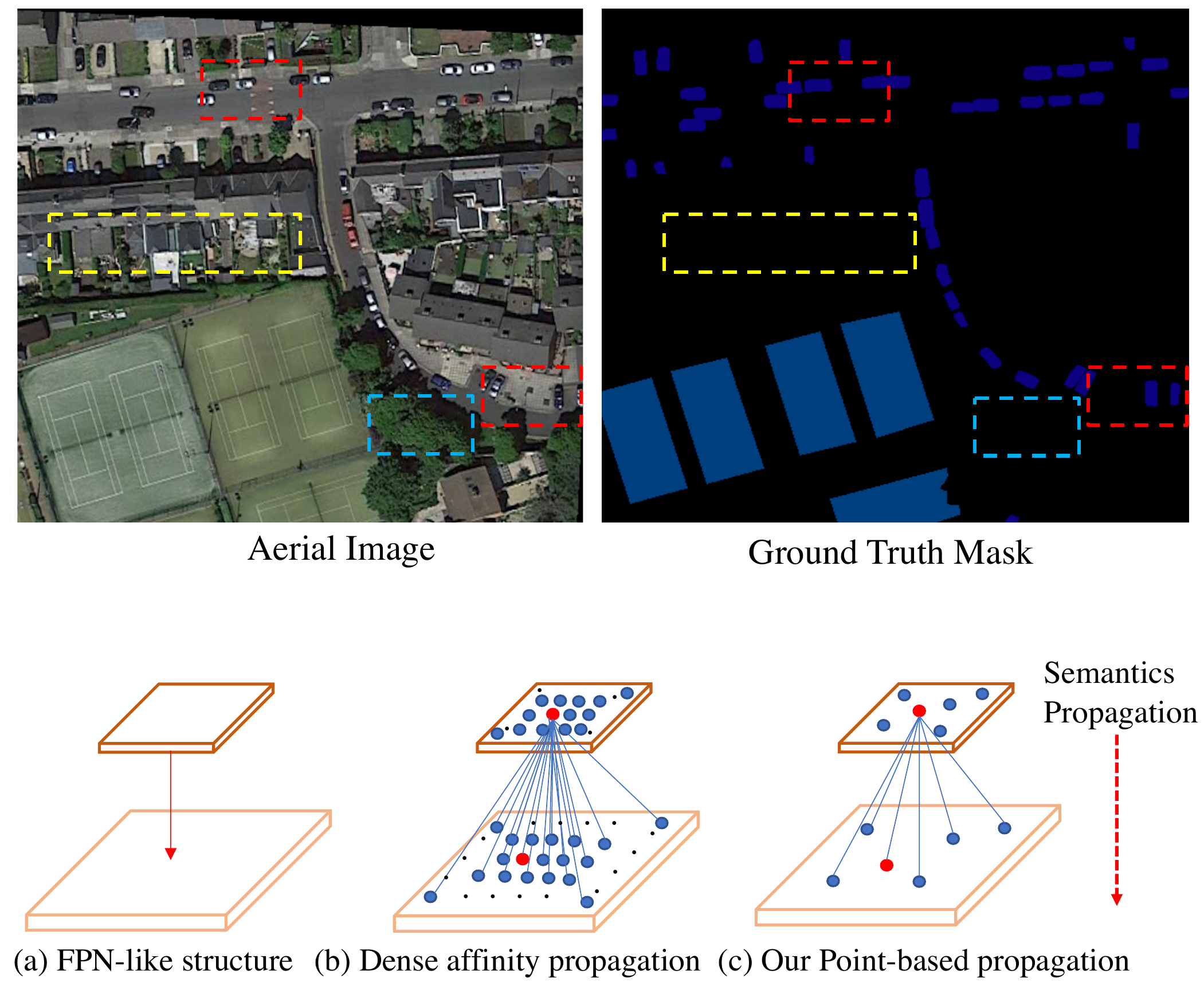}
	\caption{
	\small Illustration of an aerial image segmentation example and our proposed module. The first row presents the input image and ground truth with complex backgrounds and small objects. The second row indicates the schematic diagram on {\it dense affinity propagation} and our proposed {\it point-based propagation} module.
	}
	\label{fig:teaser}
\end{figure}

However, besides the large scale variation problems in most semantic segmentation datasets~\cite{Cityscapes,ADE20K,pcontext-data,coco_stuff}, aerial images have their own challenging problems including high background complexity~\cite{farSeg_iSAID}, background and foreground imbalance~\cite{iSAID}, tiny foreground objects in high resolution images. As shown in the first row of Fig.\ref{fig:teaser}, the red boxes show the tiny objects in the scene while the yellow box and blue box show the complex background context, including houses and trees receptively. Current general semantic segmentation methods mainly focus on scale variation in the natural scene by building multi-scale feature representation~\cite{pspnet,deeplabv3} or enhancing the object boundaries with specific design module~\cite{kirillov2019pointrend,gated-scnn}. They fail to work well due to the lack of explicit modeling for the foreground objects. For example, several dense affinity-based methods~\cite{DAnet,ocnet} also obtain inferior results mainly because the imbalanced and complex background will fool the affinity learning on small objects. For example, both yellow boxes and blue boxes have the same semantic meaning of background but with a huge appearance change. Dense affinity learning forces pixels on small objects to absorb such noisy context which leads to inferior segmentation results. FarSeg~\cite{farSeg_iSAID} adopts FPN-like~\cite{fpn} design and solve the background and foreground imbalance problems by introducing a foreground-aware relation module. However, for small objects, there still exist some semantic gaps in different features in FPN. Namely, the gap is between the high-resolution features with low semantic information and low-resolution features with high semantic information. As shown in Fig.\ref{fig:teaser}, tiny objects like cars need more semantic information in lower layers with high resolution.

In this paper, we propose a point-based information propagation module to handle the previous problems stated above. We propose PointFlow Module (PFM), a novel and efficient module for specific semantic points propagation between adjacent features. Our module is based on the FPN framework~\cite{fpn, kirillov2019panoptic} to bridge the semantic gap. As shown in the last row of Fig.\ref{fig:teaser}, rather than simple fusion or dense affinity propagation on each point as the previous work non-local module~\cite{Nonlocal}, PointFlow selects the several representative points between any adjacent feature pyramid levels. In particular, we design Dual Point Matcher by selecting matched point features from the salient area and object boundaries receptively. The former is obtained from explicit max pooling operation on the learned salient map. The latter is conditioned on the predicted object boundaries where we adopt a subtraction-based prediction. Then the point-wise affinity is estimated according to the point features that are sampled from both adjacent features. Finally, the higher layer points are fused into lower layers according to the affinity map. Our PFMs select and propagate points on foreground objects and sampled background areas to simultaneously handle both the semantic gap and foreground-background imbalance problem. 

Then we carry out detailed studies and analysis on PFM in the experiment part, where it improves the various methods by a large margin with negligible GFlops increase. Based on the FPN framework, by inserting PFMs between feature pyramids, we propose the PFNet. In particular, PFNet surpasses the previous method FarSeg~\cite{farSeg_iSAID} by \textbf{3.2\%} point on iSAID~\cite{iSAID}. Moreover, we also benchmark the recent state-of-the-art general semantic segmentation methods~\cite{HRNet,ocnet,EMAnet} on three aerial segmentation datasets including iSAID, Vaihingen and Postdam for the community. Benefited from efficient FPN design~\cite{kirillov2019panoptic}, our PFNet also achieves the best speed and accuracy on three benchmarks. Finally, we further verify the effectiveness of PFM on general semantic segmentation benchmarks, including Cityscapes~\cite{Cityscapes}, ADE-20k~\cite{ADE20K}, and BDD~\cite{yu2020bdd100k} and it achieves considerable results with previous work~\cite{deeplabv3,ocnet} with fewer GFLOPs. Our main contributions are three-fold:

1) We propose PointFlow Module (PFM), a novel and efficient module for poise-wised affinity learning, and we design a Dual Point Matcher to select the matched sparse points from salient areas and boundaries in a complementary manner.

2) We append PFM into FPN architecture and build a pyramid propagation network called PFNet.

3) Extensive experiments and analysis indicate the efficacy of PFM. We benchmark 15 state-of-the-art general segmentation methods on three aerial benchmarks. Our PFNet achieves state-of-the-art results on those benchmarks also with the best speed and accuracy trade-off. We further prove the generality of our method on three general semantic segmentation datasets.

%% file: 2relatedwork.tex
\section{Related Work}
\noindent
\textbf{General Semantic Segmentation}
The general semantic segmentation has been eminently motivated by the fully-convolutional networks (FCNs)~\cite{fcn}. The following works~\cite{pspnet,deeplabv2,deeplabv3,deeplabv3p,denseaspp,HRNet,li2019gald} mainly exploit the spatial context to overcome the limited receptive field of convolution layer which leads to the multi-scale feature representation. For example, ASPP~\cite{deeplabv3} utilizes atrous convolutions~\cite{dilation} with different atrous rate to extract features with the different receptive field, while PPM~\cite{pspnet} generates pyramidal feature maps via pyramid pooling. Several work~\cite{unet,segnet,upernet,Exfuse,densedecoder,kirillov2019panoptic,xiangtl_gff} use the encoder-decoder architecture to refine the output details. Recent works~\cite{EMAnet,ocnet,ccnet,ACFNet,CoCurrentNet,annet,OCRnet,glore_gcn,beyond_grids,zhang2019dual,Zhang_2020_CVPR_dgmn,Li2020SRNet,yu2020representativeGNN,li2020spatial} propose to use non-local-like operators or losses ~\cite{transformer,non_local,Li_2021_CVPR,yu2020context_prior} to harvest the global context of input images. Meanwhile, several works~\cite{kirillov2019pointrend,gated-scnn,yuan2020segfix,li2020improving} propose to refine the object boundaries via specific designed processing. These general semantic segmentation methods ignore the special issues including imbalanced foreground-background pixels for modeling the context and increased small foreground objects in the Aerial Imagery. Thus these methods get inferior results which will be shown in the next section.

\noindent
\textbf{Semantic Segmentation of Aerial Imagery}
Several earlier works~\cite{kaiser2017learning_aerial,marmanis2018classification,marcos2018land} focus on using multi-level semantic features on local patterns of images using deep CNN. Also, there exist a lot of applications, such as land use~\cite{huang2018urban}, building or road extraction~\cite{building_extraction,road_boundary_extraction,xu2018building_extraction,bastani2018roadtracer}, agriculture vision~\cite{agriculture}. They design specific methods based on existing semantic segmentation methods for special application scenario. In particular, relation net~\cite{relation_aerial} captures long-range spatial relationships between entities by proposing spatial and channel relation modules. Recently, FarSeg~\cite{farSeg_iSAID} propose relation-based and optimization-based foreground modeling to handle the foreground-background imbalance problems in remote sensing imagery. However, the missing explicit exploration of semantics propagation between adjacent features limits the performance on the segmentation of small objects.

\noindent
\textbf{Multi Scale Feature Fusion}
Based on the FPN framework~\cite{fpn}, rather than simple top-down additional fusion, several works propose to fuse feature through gates~\cite{ding2018context,xiangtl_gff}, neural architecture search~\cite{nasfpn}, pixel-level alignment~\cite{sfnet} or adding bottom up path~\cite{PANet}, dense affinity learning propagation\cite{feature_pyramid_transformer}.
Such full fusion methods may emphasize background objects like roads where the imbalance problem exists widely in aerial images. Our proposed PFM follows the design of FPN by propagating the semantics from the top to bottom. In contrast, rather than full fusion like previous works, our methods are based on point-level which select the several representative points to overcome the pixel imbalance problems in aerial imagery and leads to better results.

%% file: 3method.tex
\section{Method}
In this section, we will first introduce some potential issues on dense point affinity learning for aerial segmentation task. Then we will provide detailed descriptions of our PointFlow module(PFM) to resolve the issues by selecting key semantic points for propagation efficiently. Finally, we will present our PFNet for aerial imagery segmentation.

\begin{table}[!t]\setlength{\tabcolsep}{6pt}
	\centering
	\begin{threeparttable}
		\scalebox{0.75}{
			\begin{tabular}{l c c c }
				\toprule[0.15em]
		Method	& OS & mIoU  & $\Delta$ \\
				\toprule[0.15em]
	    dialated FCN\cite{fcn,dilation}(baseline) & 8 & 59.0 & -\\
		DAnet\cite{pspnet}  & 8& 30.3 & 28.7$\downarrow$\\
        OCnet(ASP-OC)~\cite{ocnet} & 8 &  40.2 & 18.8$\downarrow$ \\
        DAnet+FPN~\cite{fpn} & 8& 59.3 & 0.3$\uparrow$\\
        DAnet+our PFNet decoder & 8&  65.6 & 6.6$\uparrow$\\
        \hline
        SemanticFPN~\cite{kirillov2019panoptic}(baseline) & 32 & 61.3 & -\\
        +dense affinity~\cite{feature_pyramid_transformer}& 32 & 58.9& 2.4$\downarrow$\\
        +our PFM & 32 & 65.0 & 3.7$\uparrow$ \\
	\bottomrule[0.1em]
	\end{tabular}}
		\caption{ \small Simple experiment results on iSAID validation dataset. The dense affinity results in inferior results over various baselines. Appending our proposed PointFlow module results in a significant gain. OS: Output Stride in backbone.
		}
		\label{tab:results_on_iSAID_datasets_dense_affinity_methods}
	\end{threeparttable}
\end{table}

\subsection{Preliminary}
Recent dense affinity based methods~\cite{Nonlocal,ocnet,DAnet,transformer,feature_pyramid_transformer} have shown progressive results for semantic segmentation. The core idea of these methods is to model the pixel-wised relationship to harvesting the global context. As shown in Equ.~\ref{equ:dense_affinity}, in the view of self-attention~\cite{transformer}, each pixel $p$ in 2-D input feature $F$ $\in \RR^{C \times H \times W}$ is connected to all the other pixels to calculate the pixel-wised affinity where $A$ is the affinity function and it outputs affinity matrix $\in \RR^{HW \times HW}$. $C$, $H$, and $W$ denote the channel dimension, height, and width, respectively. Note that definitions of $A$ can be different; we use the same label for simplicity. 

\begin{equation}
    F^{r}(p) =  A(F(p), F(p)) {F(p)}
    \label{equ:dense_affinity}
\end{equation}

However, applying these methods directly on the iSAID dataset leads to inferior results even compared with various baseline methods, as shown in Tab.~\ref{tab:results_on_iSAID_datasets_dense_affinity_methods} whether such module is appended after FCN backbone or is inserted into feature pyramids. The reason has two folds: (1) There exist extremely imbalanced foreground-background objects in the iSAID dataset. Explicit affinity modeling on complex background brings noise for outputs. (2) Too many small objects exist on the iSAID dataset, which requires high resolution and high semantic representation.

To solve the first problem, rather than dense affinity modeling, we can use a point sampler $\beta$ to select matched representative points $\hat{p}$ to balance the background context ratio while keeping efficiency. For the second problem, to fill the semantic gap on small objects, we adopt the FPN framework and change the inputs of $A$ by using adjacent features in a top-down manner shown in Equ.~\ref{equ:point_affinity}:
\begin{equation}
    F^{r}(\hat{p}) =  A(\beta(F_{l}(\hat{p})), \beta(F_{l-1}(\hat{p}))) \beta(F_{l}(\hat{p})))
    \label{equ:point_affinity}
\end{equation}
where $F_{l}$ and $F_{l-1}$ are adjacent features in the FPN framework and $\hat{p}$ is sampled pixels for affinity modeling. We will detail the $\beta$ in the following part. As shown in Tab.\ref{tab:results_on_iSAID_datasets_dense_affinity_methods}, our method improves the baselines by a significant margin.

\subsection{PointFlow Module}

\begin{figure*}
	\centering
	\includegraphics[width=1.0\linewidth]{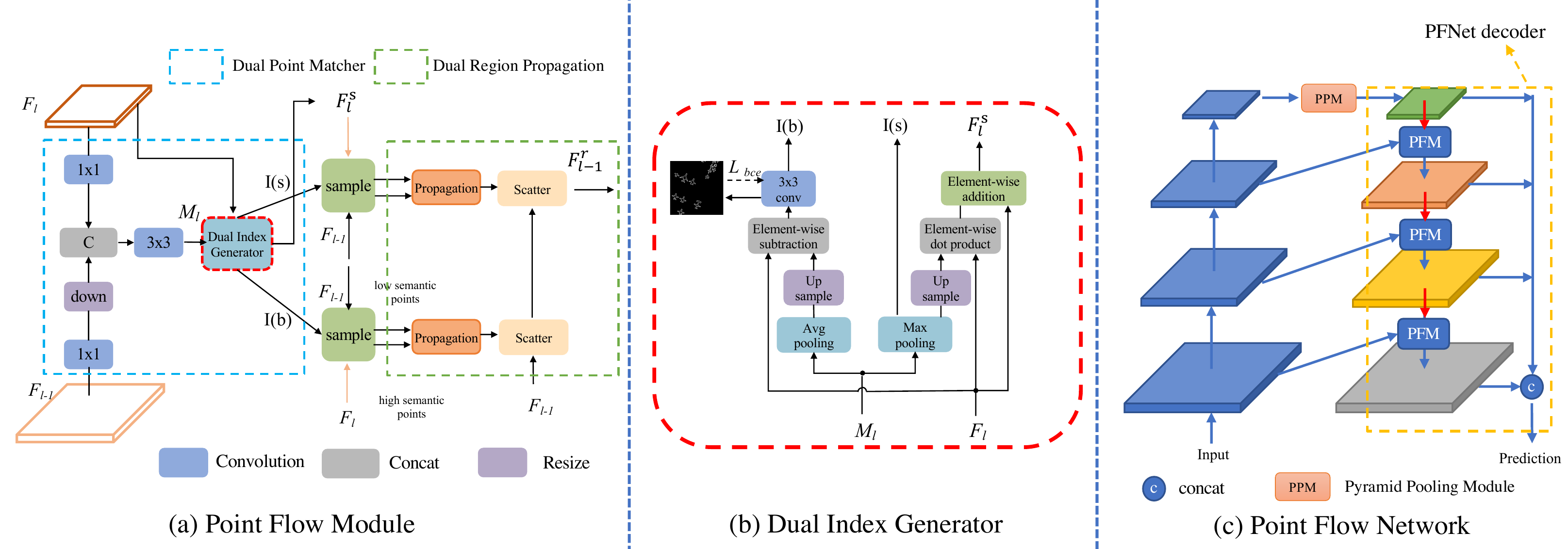}
	\caption{
	(a), The overal pipeline of our proposed PointFlow Module. Left: Two adjacent features with one salient map are sent to the Dual Index Generator to obtain the sampled indexes. Right: The sampled point features are propagated from top to the bottom and finally scattered into the low level features point-wisely.
	(b), The detailed operation on proposed Dual Index Generator.
	(c), We design the PF Network Architecture by inserting PF modules into FPN-like framework.
	}
	\label{fig:network}
\end{figure*}

\noindent \textbf{Motivation and Overview} As the previous section shows the limitation of dense affinity on aerial image segmentation, we argue that unnecessary background pixels context may bring noises for foreground objects. Considering this, we propose to propagate context information through selective points, which can keep the efficiency in both speed and memory. Meanwhile, the semantic gap problems can also be fixed after propagation leading to high-resolution feature representation with high semantics, which is why we adopt FPN-framework design~\cite{fpn} in a top-down manner. Since our framework works in a top-down manner, and the semantics flow into low-level features through points, we name our module PointFlow. Our PointFlow is built on the FPN framework~\cite{fpn}, where the feature map of each level is compressed into the same channel depth through two 1$\times$1 convolution layers before entering the next level. Our module takes two adjacent feature maps as inputs $F_{l-1} \in \RR^{C \times H \times W}$ and $F_{l} \in \RR^{C\times H/2 \times W/2}$ as the inputs where $l$ means the index of feature pyramid and output refined $F_{l-1}^{r} \in \RR^{C \times H \times W}$. For modeling $\beta$, we propose the Dual Point Matcher to select the points, and then the point-wise affinity can be calculated between adjacent points. Finally, the points with high-resolution and low semantics can be enhanced by the points with low-resolution high semantics according to the estimated affinity map. The process is shown in Fig.~\ref{fig:network}(a).

\noindent \textbf{Dual Point Matcher} The critical issue is how to find the corresponding points between two adjacent maps. We argue that most salient areas can be represented as key points for balanced pixel-level propagation due to the unbalanced pixels between the foreground and background. Meanwhile, since there are many small objects in aerial scenes that need more fine-grained location cues, the boundary areas can also be considered the key points. Thus we design a novel Dual Point Matcher to consider the most salient part of inputs and object boundaries at the same time.  The Dual Point Matcher has two steps: (1) Generate the salient map. (2) Generate sampled indexes from Dual Index Generator. 

For the first step, we combine the input feature maps where the high-resolution part $F_{l-1}$ is downsampled into the same low resolution through bilinear interpolation. The resized feature is denoted as $\widetilde F_{l-1}$. Then we perform one $3 \times 3$ convolution following with sigmoid function to generate the saliency map $M_{l}$. The process is shown as follows: 
\begin{equation}
    M_{l} = Sigmoid(\text{conv}_l(\text{Concat}(F_{l},  \widetilde F_{l-1}))),
    \label{equ:salient_points_sample}
\end{equation}
For the second step, we take $F_{l}$ and $M_{l}$ as the inputs of the Dual Index Generator. We perform the adaptive max pooling on such map to obtain the most salient points. To highlight the salient part of foreground objects, we multiply such map on $F_{l}$ with residual design as attention map shown in Equ.~\ref{equ:salient_points_attention}:
\begin{equation}
    F_{l}^{s} = MaxPool(M_{l}) \times F_{l} + F_{l},
    \label{equ:salient_points_attention}
\end{equation}
We simply choose the salient indexes from $MaxPool(M_{l})$. $K$ is the number of pooled points, and it equals to the product of adaptive pooling kernels. We denote the salient indexes as $I(s)$ for short. 

For boundary point selection, rather than simply using the binary supervision on the input feature $F_{l}$ or $F_{l-1}$ for boundary prediction, we propose to adopt residual prediction on the $F_{l}$. Our method is motivated by Laplacian pyramids in image processing~\cite{adelson1984pyramid,burt1981fast_pyramid}. In Laplacian pyramids, the edge part of original images can be obtained by subtracting the smoothed upsampled images. Motivated by that, we use the average pooling on saliency map $M_{l}$ and multiply the pooled map on $F_{l}$ for smoothing inner content, then we subtract the such smoothed part from $F_{l}$ to generate the sharpened feature $\widetilde F_{l}^{b}$ for boundary prediction. The process is shown in Equ.~\ref{equ:boundary_point_matcher}:
\begin{equation}
    \widetilde F_{l}^{b} =  F_{l} - AvgPool(M_{l}) \times F_{l},
    \label{equ:boundary_point_matcher}
\end{equation}
After the boundary prediction using $\widetilde F_{l}^{b}$, we obtain the boundary map $B_{l}$. Following the previous step, we simply sample Top-K points from the edge maps(K=128 by experiment) according to their confidence scores. We denote the boundary indices $I(b)$ for short. In total, the Dual Index Generator samples the key points in an orthogonal way by selecting points from specific regions according to the salient map $M_{l}$. The total process of Dual Index Generator is shown in Fig.~\ref{fig:network}(b).

\noindent \textbf{Dual Region Propagation} After the point matcher, we obtain the indexes $I(s)$ and $I(b)$, respectively. Then we sample the points from map from salient feature $F_{l}^{s} $ and original input feature $F_{l-1}$. For each selected point, a point-wise feature representation is extracted on both adjacent input features. Note that features $f$ for a real-value point are computed by bilinear interpolation of 4 nearest neighbors that are on the regular grid. We use normalized girds during the implementation. We denote $f_{l}^{s}$ and $f_{l}^{b}$ as sampled feature point at stage $l$ for salient part and boundary part. We propagate those sampled points independently. For each sampled point $\hat{p}$, the top-down propagation process is shown in Equ~\ref{equ:pointflow_formulation}. 
\begin{equation}
     f_{l-1}(\hat{p})^{r} = \sum_{i\in{\{I(b),I(s)\}}} A(f_{l-1}^{i}(\hat{p}), f_{l}^{i}(\hat{p})) f_{l}^{i}(\hat{p}) + f_{l-1}^{i}(\hat{p}) ,
    \label{equ:pointflow_formulation}
\end{equation}
where $A$ is affinity function, $i$ means the indexes whether from $I(s)$ or $I(b)$. For $A$, we use the point-wise matrix multiplication along with softmax function for normalization. Following the previous work~\cite{resnet}, we adopt the residual design for easier training. 
We calculate the sampled high semantic points through point-wise affinity according to the semantic similarity on sampled points with low semantics, which avoids the redundant background information in the aerial scene. Since we propagate semantics two times independently, we term two flows as {\it salient point flow} and {\it boundary point flow}, respectively. Finally, the refined feature $F_{l-1}^{r} $ is obtained by scattering the $f_{l-1}^{r}$ into $F_{l-1}$ according to the indices $I(s)$ and $I(b)$.  

\subsection{Network Architecture}
\noindent \textbf{Overview} Fig.~\ref{fig:network} illustrates the our network architecture, which contains a bottom-up pathway as the encoder and a top-down pathway as the decoder. The encoder is backbone network with multiple feature pyramid outputs while the decoder is lightweight FPN equipped with our PFMs.

\noindent \textbf{Network Architecture} The encoder uses the ImageNet pretrained backbone with OS 32 rather dilation stragety with OS 8 for efficient inference. 
We additionally adopt the Pyramid Pooling Module (PPM)~\cite{pspnet} for its superior efficiency and effectiveness to capture contextual information. In our setting, the output of PPM has the same resolution as that of the last stage. PFNet decoder takes feature maps from the encoder and uses the refined feature pyramid for final aerial segmentation according to previous work design~\cite{farSeg_iSAID,kirillov2019panoptic}. By simply replacing normal bilinear up-sampling with our PF module in top-down pathway of FPN, the PFNet decoder finally concatenates all the refined $F_{l}^{r}$(where $l$ ranges from 2 to 5) by upsampling the inputs to the same resolution(1/4 resolution of input) and perform prediction. Note that our module can also be integrated into other architectures including Deeplabv3~\cite{deeplabv3} with a slight modification by appending such decoder after its head. More details can be found in the experiment part.

\noindent \textbf{Loss Function} For edge prediction in each PFM, we adopt binary BCE loss $L_{bce}$. For final segmentation prediction, we adopt the cross-entropy loss. The two losses are weighted to 1 by default.

%% file: 4experiment.tex
%


\section{Experiments}
\noindent \textbf{Overview:} We will firstly perform ablation studies on iSAID dataset and give detailed analysis and comparison on PFM. Then we benchmark several recent works on Vaihingen and Potsdam datasets. Finally, we prove the generalizability of our module on general segmentation datasets.

\subsection{Aerial Image Segmentation}
\noindent \textbf{DataSets:} We use iSAID~\cite{iSAID} dataset for ablation studies and report results on remaining datasets. iSAID~\cite{iSAID} consists of 2,806 HSR images. The iSAID dataset provides 655,451 instances annotations over 15 categories of the object and it is the largest dataset for instance segmentation in the HSR remote sensing imagery. We also use Vaihingen and Postdam datasets\footnote{https://www2.isprs.org/commissions/comm2/wg4/benchmark/} for benchmarking.

\noindent \textbf{Implementation detail and Metrics:} We adopt ResNet-50~\cite{resnet} by default. Following the same setting~\cite{farSeg_iSAID}, for all the experiments, these models are trained with 16 epoch on cropped images. For data augmentation, horizontal and vertical flip, rotation of 90 · k (k = 1, 2, 3) degree were adopted during training. For data preprocessing, we crop the image into a fixed size of (896, 896) using a sliding window striding 512 pixels. We use the mean intersection over union (mIoU) as the main metric for object segmentation to evaluate the proposed method if not specified. The baseline for ablation studies is Semantic-FPN~\cite{kirillov2019panoptic} with OS 32.

\begin{table*}[t]
	\begin{minipage}[!t]{\linewidth}
	    \centering
		\begin{minipage}{.38\linewidth}
		\subfloat[Effect of dual flow propagation on baseline.]{
		\resizebox{1\textwidth}{!}{%
		\centering
		\begin{tabular}{ c c c l l }
					\toprule[0.15em]
					  +PPM & +salient point flow & + boundary point flow & mIoU(\%)  \\  
					\toprule[0.15em]
				     - &  - & - & 61.3 \\
					 \checkmark & - & - & 63.8 \\
					 \- & \checkmark & \checkmark & 65.0  \\
					\checkmark &\checkmark & - &   64.8 \\
					\checkmark & -  & \checkmark &  66.2  \\
					\checkmark &\checkmark &\checkmark & 66.9 \\
					\hline
				\end{tabular}}}\hspace{3mm}
	    \end{minipage}
	    \hspace{2mm}
	    \begin{minipage}{.32\linewidth}
		\subfloat[Effect of Insertion Position. $\hat{\FF_l}$ means the position between $\FF_l$ and $\FF_l-1$. ]{
		\resizebox{0.85\textwidth}{!}{%
		\centering
		\begin{tabular}{c c c c l}
			\toprule[0.15em]
			Method & $\hat{\FF_3}$ & $\hat{\FF_4}$ & $\hat{\FF_5}$ & mIoU(\%) \\  
		    \toprule[0.15em]
			Baseline+PPM &  &  &  & 63.8 \\ 
			& \checkmark & & & 65.8 \\
			& & \checkmark & & 65.6  \\
			& & & \checkmark & 65.5  \\
			\midrule
			& &\checkmark &\checkmark & 66.5 \\
			&\checkmark &\checkmark &\checkmark & 66.9 \\
			\hline
		\end{tabular}	
				}}\hspace{3mm}
		\end{minipage}
		\hspace{2mm}
		\begin{minipage}{.24\linewidth}
		\subfloat[Comparison with Other Propagation Methods.]{
			\footnotesize
			\resizebox{0.95\textwidth}{!}{%
			\begin{tabular}{l|c}
				\toprule[0.15em]
				Settings & mIoU(\%)   \\
				\toprule[0.15em]
				baseline+PPM &  63.8 \\
				\hline 
			    +DCNv1~\cite{deformable} & 65.2 \\
			    \hline
			    +DCNv2~\cite{deformablev2} & 65.6 \\
				\hline
				+desne affinity flow~\cite{non_local} & 62.0 \\
				\hline
				+FAM~\cite{sfnet} & 65.7 \\
				\hline
				+ Ours & 66.9 \\
 				\hline
		\end{tabular}}}
		\end{minipage}
	\end{minipage}
    \vspace{3pt}
	\begin{minipage}[!t]{\linewidth}
	   \centering
		\begin{minipage}{.24\linewidth}
		\subfloat[Effect of salient point sampling in Dual Index Generator. \label{expr:ablation_norm}]{
			\footnotesize
			\resizebox{0.85\textwidth}{!}{%
			\begin{tabular}{l|c }
				\toprule[0.15em]
				Sampling Method & mIoU(\%)  \\
				\toprule[0.15em]
				baseline+PPM & 63.8 \\
			    uniform random & 64.0 \\
			    attention based & 64.2  \\
			    Our max pooling & 64.8 \\
				\hline
		\end{tabular}}}
		\end{minipage}
		\begin{minipage}{.20\linewidth}
		\subfloat[Effect of propagation direction.]{
			\footnotesize
			\resizebox{0.85\textwidth}{!}{%
			\begin{tabular}{l|c}
				\toprule[0.15em]
				Settings & mIoU(\%)   \\
				\toprule[0.15em]
				baseline + PPM & 63.8 \\
				\hline 
			    top-down(td) &  66.9 \\
				\hline
				bottom-up(bu) & 47.3 \\
				\hline
				td then bu & 54.5 \\
				\hline
		\end{tabular}}}
		\end{minipage}
        \begin{minipage}{.20\linewidth}
		\subfloat[Effect of edge generation module in Dual Index Generator.]{
			\footnotesize
			\resizebox{0.95\textwidth}{!}{%
			\begin{tabular}{l|c }
				\toprule[0.15em]
				Settings  & mIoU(\%)   \\
				\toprule[0.15em]
				baseline+PPM & 63.8 \\
				\hline
                direct prediction & 65.7 \\
				\hline
				addition prediction  &  65.5 \\
				\hline
				Our subtraction based & 66.2 \\
				\hline
		\end{tabular}}}
	    \end{minipage}
	    \begin{minipage}{.32\linewidth}
		\subfloat[Application on Other Architectures. \label{expr:ablation_placement}]{
			\footnotesize
			\resizebox{0.85\textwidth}{!}{%
			\begin{tabular}{l|c c c}
				\toprule[0.20em]
				Network  & Backbone & mIoU(\%) & GFlops \\
				\toprule[0.20em]
				Deeplabv3~\cite{deeplabv3} & ResNet50 & 60.4 & 168.4 \\
				Deeplabv3~\cite{deeplabv3} & ResNet101 & 61.5 & 264.1 \\
				\hline
				+FPN & ResNet50 & 62.3 & 183.4 \\
				+PF decoder & ResNet50 & 65.6 & 185.2 \\
				\hline
				CCNet~\cite{ccnet} & ResNet50 & 58.3 & 206.5 \\
				\hline
				+FPN & ResNet50& 60.2 & 220.8 \\
				+PF decoder & ResNet50 &  65.3 & 223.2 \\
				\hline
		\end{tabular}}}
		\end{minipage}
	\end{minipage}
	
	\caption{\textbf{Ablation studies.} We first verify the effect of each module and comparison results in the first row. Then we verify several design choices and generality of our module in the second row.}\label{tab:ablations}
\end{table*}

\noindent \textbf{Effectiveness on baseline models:} In Tab.~\ref{tab:ablations}(a), adopting our PFMs leads to better results than appending PPM~\cite{pspnet} shown in both 2nd and 3rd rows with about 1.2 \% gap. After applying both PPM and PFM, there is a significant gain over the baseline models shown in the last row. Only applying boundary flow is slightly better than applying salient point flow which indicates the small object problems are more severe than foreground-background imbalance problems in this dataset. In Tab.~\ref{tab:ablations} (b), we explore the effect on insertion position with our PFMs. From the first three rows, PF improves all stages and gets the greatest improvement at the first stage, which shows that the semantic gap is more severe for small objects in lower layers. After appending all FPMs, we achieve the best result shown in the last row.

\noindent \textbf{Comparison with feature fusion methods:}
Tab.~\ref{tab:ablations}(c) give several feature fusion methods~\cite{deformable, sfnet, feature_pyramid_transformer} used on scene understanding tasks. For all the methods, we replace these modules into the same position on PFnet decoder as in Fig~\ref{fig:network}(c) for fair comparison. Compared with DCN-like methods~\cite{deformable,deformablev2,sfnet}, our method leads to significant gain over them since our method can better handle the foreground semantics propagation.

\noindent \textbf{Ablation on design choices:} We give more detailed design studies in the second row of Tab.~\ref{tab:ablations}. Tab.~\ref{tab:ablations}(d) explores several sampling methods for salient points sampling. Attention based method is directly selecting top-K(K=128) points from $M_{l}$ while uniform random sample is done by randomly selecting one pixels from $7\times7$ neighbor region of $M_{l}$(We report average result of 10 times experiments). Our max pooling based methods work the best among them. Tab.~\ref{tab:ablations}(e) shows the propagation direction of PFM. Adding bottom up fusing leads to bad results mainly because more background context is introduced into the head which verifies our motivation of flowing semantics into the bottom. Tab.~\ref{tab:ablations} (f) shows the effect results on edge prediction. Our subtraction based prediction has better results mainly due to better boundary prediction. This is also verified in Tab.~\ref{tab:results_semantic_boundaries}. 

\begin{figure}
	\centering
	\includegraphics[width=\linewidth]{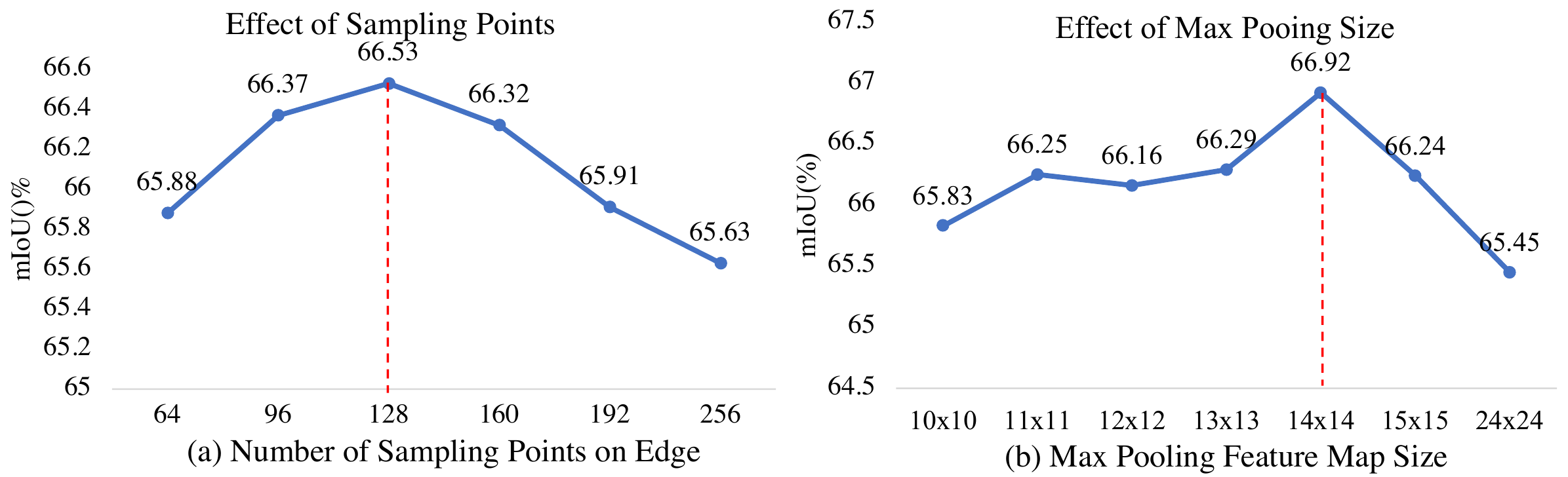}
	\caption{
	\small Ablation studies on the number of sampled point for both point flows. Best view it in color and zoom in.
	}
	\label{fig:sampled_points}
\end{figure}

\noindent \textbf{Ablation on Number of Sampled Points:} We first verify the best number of sampled points on boundary point flow in Fig.~\ref{fig:sampled_points}(a) by increasing the number of sampled pixels where we find the best number is 128. Sampling more points leads to inferior results which indicates missing background context is also important. Appending the boundary flow as the strong baseline, we explore the kernel size of salient point flow where we find the best kernel size 14 $\times$ 14 (256 points in total) in Fig.~\ref{fig:sampled_points}(b). After selecting more points (24$\times$24, 576 points in total), the performance drops a lot since the imbalance problems exist. This verifies the same conclusion that the dense affinity leads to bad results.

\noindent \textbf{Application on Various Methods:} Our PFM can be easily adopted into several existing networks by extending PFNet decoder (shown in Fig.~\ref{fig:network}(c) yellow box) after their heads. More details can be referred to supplementary. In Tab.~\ref{tab:ablations}(g), we verify two work including Deeplabv3\cite{deeplabv3} and CCNet~\cite{ccnet} where we obtain significant gains over these baselines. This proves the generalization of our methods. Our method outperforms ResNet101-based models which indicates the improvement is not obtained by extra parameters introduced by PFM.

\begin{table}[!t]\setlength{\tabcolsep}{6pt}
	\centering
	\begin{threeparttable}
		\scalebox{0.75}{
			\begin{tabular}{l c c c c c c }
				\toprule[0.15em]
				Method    & mIoU  & F1(12px) & F(9px) & F1(5px) & F1(3px)  \\
				\toprule[0.15em]
		baseline+PPM  &   63.8 & 88.2 & 86.2 & 85.6 & 84.3 \\
    	+salient point flow:  &  64.8 & 88.9 & 88.1  & 87.0 & 85.4 \\
        +boundary point flow   &  66.2 & 93.2 & 91.2 & 89.0 & 88.4 \\
        +both  & 66.9 & 94.2 & 93.2 & 90.2 & 89.0 \\
        \hline
        direct prediction  & 65.7 & 89.6 & 87.5 & 86.4 & 85.8 \\
        subtraction prediction  &  66.2 & 93.2 & 91.2 & 89.0 & 88.4 \\
	\bottomrule[0.1em]
	\end{tabular}}
		\caption{Ablation study on semantic boundaries where we adopt 4 different thresholds for evaluation.
		}
		\label{tab:results_semantic_boundaries}
	\end{threeparttable}
\end{table}

\noindent \textbf{Effectiveness on Segmentation Boundaries:} We further verify the boundary improvements using F1-score metric~\cite{VOS_benchmark} with different pixel thresholds in Tab.~\ref{tab:results_semantic_boundaries}. Appending boundary point flow leads to more significant improvements than salient point flow due to the explicit supervision and propagation on boundary pixels. Adopting both flows leads to the best results and it indicates the complemented property of our approach. Moreover, as shown in the last row of Tab.~\ref{tab:results_semantic_boundaries}, our subtraction based edge prediction results better than direct prediction where it has better mask boundary. We include boundary prediction results in supplementary.

\begin{figure}
	\centering
	\includegraphics[width=0.95\linewidth]{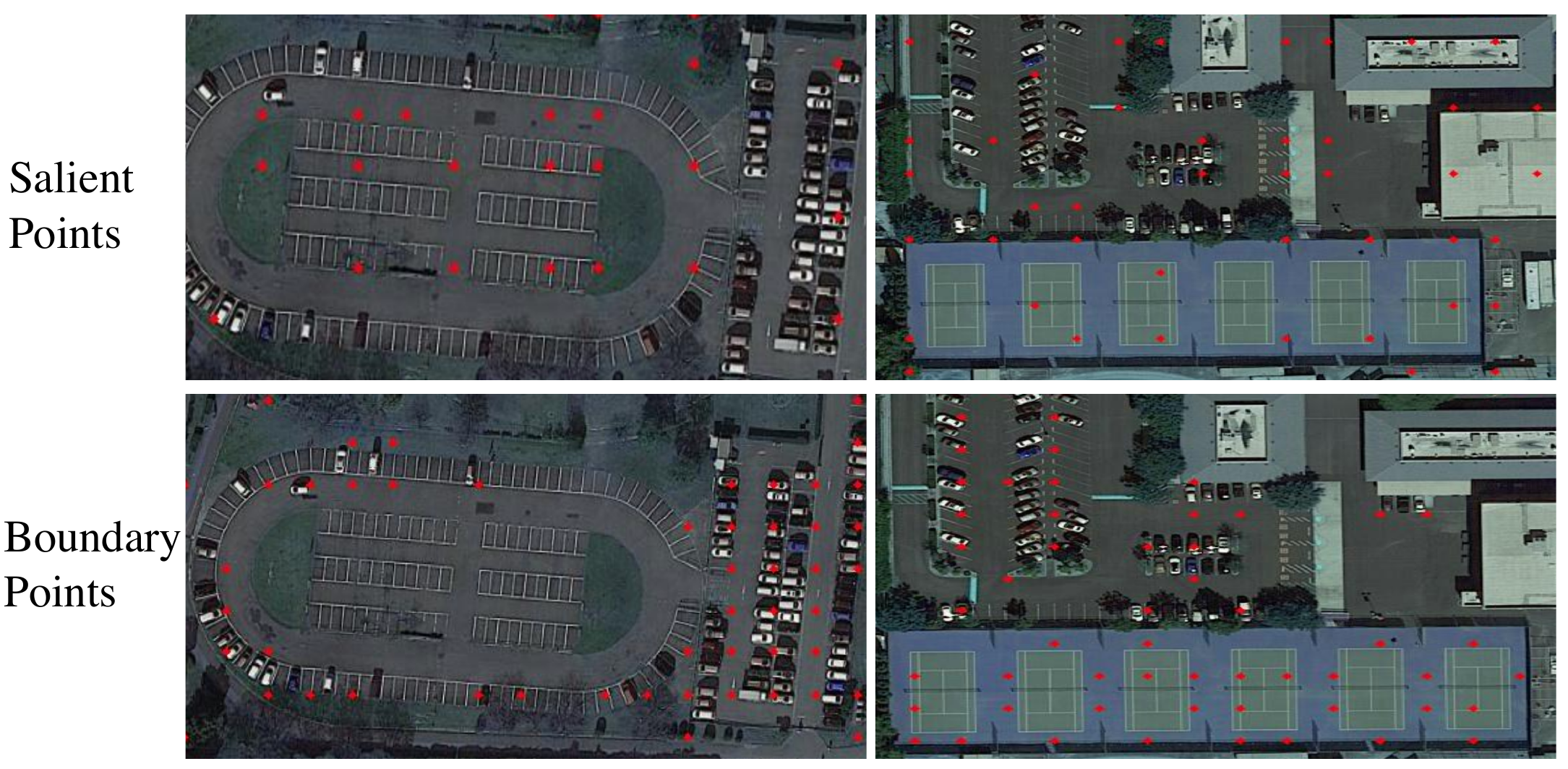}
	\caption{
	\small Visualization of sampled points for both point flows. Top: Salient Flow points. Bottom: Boundary Flow points. Best view it on screen.
	}
	\label{fig:vis_sampled_points}
\end{figure}

\noindent \textbf{Balanced foreground-background Points:} We analyze the ratio of sampled points on fore-ground parts over total sampled points by adding all three PFMs using validation images. Compared with baseline unbalanced points with 2.89\% computed by ground truth mask, our method improves the ratio on foreground to 7.83\% during the inference which resolves the problems of imbalanced points aggregation. We use the best model found in ablation part.

\noindent \textbf{Visualization of Sampled Points:} 
In Fig.~\ref{fig:vis_sampled_points}, we show several visual examples on sampled points on the original images. The first row gives the salient point results while the second row shows the boundary point results. We visualize the points from the PFM in the last stage. As shown in Fig.~\ref{fig:vis_sampled_points}, the salient points uniformly locate around the foreground objects and several of them are on the background sparsely. The boundary points are mainly on the boundary of large foreground objects and the inner regions on small objects because the downsampled feature representation makes it hard to predict small object boundaries. More visual examples can be found in supplementary.

\noindent \textbf{Benchmarking recent works on aerial images datasets:} Recent work FarSeg~\cite{farSeg_iSAID} reports results of several segmentation methods~\cite{pspnet,kirillov2019panoptic,deeplabv3} on iSAID datasets. We extend more representative work~\cite{ccnet,DAnet,EMAnet,kirillov2019pointrend} on iSAID, Vaihingen and Postdam datasets under the same experiment setting. Note that, for all methods, we use ResNet50 as backbone for fair comparison except for HRNet~\cite{HRNet}. The work~\cite{relation_aerial} also reports results on Vaihingen and Postdam using weak VGG-backbone~\cite{vgg}. Due to the lack of comparison with recent work, we re-implement this method using ResNet50 backbone and trained on larger cropped images and report mIoU as metric.\footnote{We will opensource all the models and code for further research.} All the methods use the single scale inference on cropped images for testing.

\begin{table}[!t]\setlength{\tabcolsep}{6pt}
	\centering
	\begin{threeparttable}
		\scalebox{0.60}{
			\begin{tabular}{l c c c }
				\toprule[0.15em]
			Method	 & Backbone & mIoU & OS \\
				\toprule[0.15em]
		DenseASPP~\cite{denseaspp} & RenNet50 & 57.3 & 8  \\
        Deeplabv3~\cite{deeplabv3} & ResNet50 & 60.4 & 8\\
        Deeplabv3+ ~\cite{deeplabv3p} & ResNet50 & 61.2 & 8\\
        RefineNet ~\cite{refinenet} & ResNet50 & 60.2  & 32 \\
		PSPNet~\cite{pspnet} & ResNet50 & 60.3 & 8\\
        OCNet-(ASP-OC)~\cite{ocnet} & ResNet50 &  40.2 & 8 \\
        EMANet~\cite{EMAnet} & ResNet50 & 55.4 & 8\\
        CCNet~\cite{ccnet} & ResNet50 & 58.3 & 8\\
        EncodingNet~\cite{context_encoding} & ResNet50 & 58.9 & 8 \\
        SemanticFPN~\cite{kirillov2019panoptic} & ResNet50 & 62.1 & 32 \\
        UPerNet~\cite{kirillov2019panoptic} & ResNet50 & 63.8 & 32 \\
        HRNet\cite{upernet} & HRNetW18 & 61.5 & 4 \\
        SFNet\cite{sfnet} & ResNet50 & 64.3 & 32 \\
        GSCNN\cite{gated-scnn} & ResNe50 & 63.4 & 8 \\
        RANet\cite{relation_aerial} & ResNet50 & 62.1 & 8 \\
        FarSeg~\cite{farSeg_iSAID} & ResNet50 & 63.7 & 32\\
        \hline
        PFNet  & ResNet50 & \bf{66.9} & 32 \\
	\bottomrule[0.1em]
	\end{tabular}}
		\caption{\small Comparison with the state-of-the-art results on iSAID dataset.
		}
		\label{tab:results_on_iSAID_datasets}
	\end{threeparttable}
\end{table}

\noindent \textbf{Comparison with the state-of-the-arts on iSAID:} 
We first benchmark more results on iSAID dataset in Tab.~\ref{tab:results_on_iSAID_datasets} and then compare our PFNet with previous work. Our PFNet achieves the state-of-the-art result among all previous work by a large margin. Our method outperforms previous state-of-the-art FarSeg~\cite{farSeg_iSAID} by 3.2\%.

\noindent \textbf{Experiments on Vaihingen and Potsdam:} Rather than the previous work~\cite{relation_aerial} cropping the images into small patches, we adopt large patches as the iSAID dataset and use more validation images for testing. That makes the segmentation more challenging. The details of train and validation splitting can be found in the supplementary.
For the Vaihingen dataset, we preprocess the images by cropping into 768$\times$ 768 patches. 
We adopt the same training setting with iSAID dataset except for 200 epochs and larger learning rate with 0.01. For the experiments on the Potsdam dataset, the images are cropped into 896$\times$896 patches. The total training epoch is set to 80 with the initial learning rate of 0.01. As shown in Tab.~\ref{tab:results_on_aerial_datasets}, we benchmark recent segmentation methods with two metrics including mIoU and mean-$F_1$. Our PFNet achieves state-of-the-art results on two benchmarks.


\begin{figure*}
	\centering
	\includegraphics[width=0.90\linewidth]{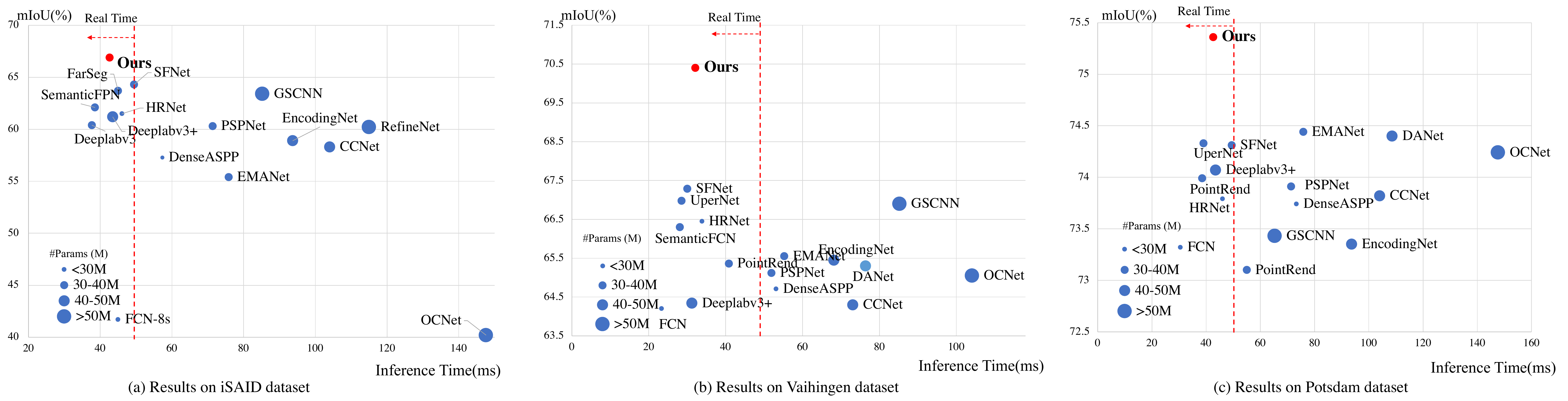}
	\caption{
	\small Speed (Inference Time) versus Accuracy (mIoU) on three aerial segmentation datasets. The radius of circles represents the number of parameters. All the methods are tested with one V-100 GPU card for fair comparison. Our PFNet achieves the best speed and accuracy trade-off on three benchmark. Real time is within 50ms. Best view it on screen and Zoom in. 
	}
	\label{fig:speed_acc}
\end{figure*}

\begin{figure*}
	\centering
	\includegraphics[width=0.90\linewidth]{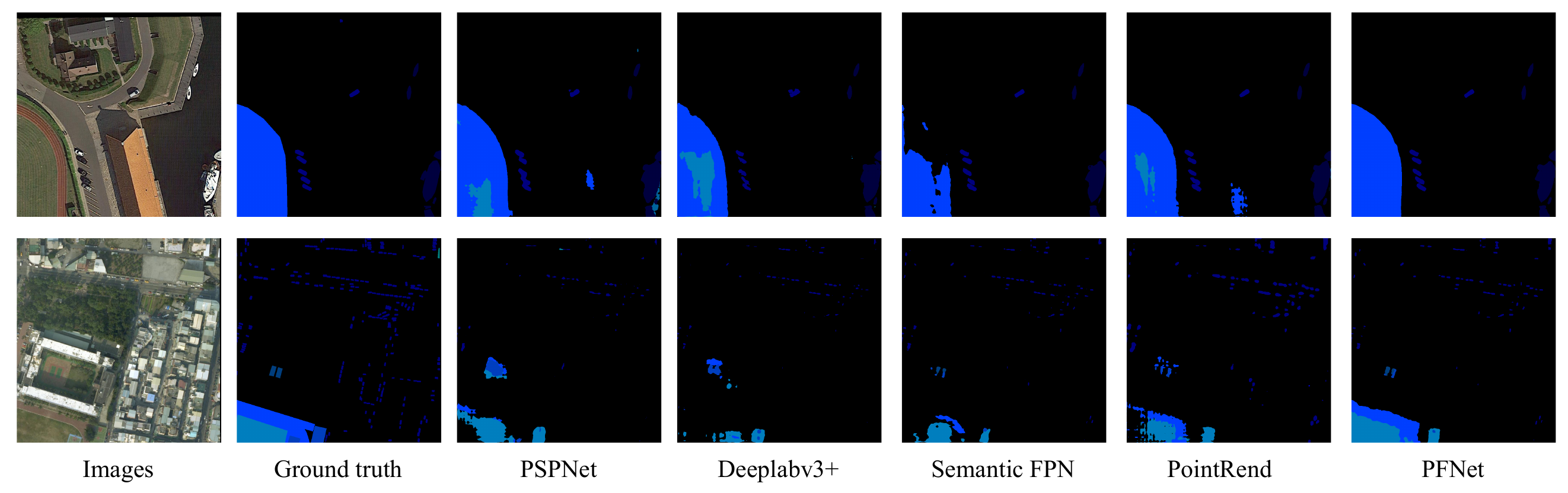}
	\caption{\small Visual results on iSAID validation set. Compared with previous works, our method obtains better segmentation results. Best view on the screen. More visual results can be found in supplementary. }
	\label{fig:vis_res_on_iSAID}
\end{figure*}

\noindent \textbf{Efficiency Comparison:} 
 In Fig.~\ref{fig:speed_acc}, we further benchmark the speed and parameters of our methods on above datasets. Compared with previous work, PFNet achieves the best speed and accuracy trade-off on those three benchmarks with fewer parameters without bells and whistles. Note that PFNet can also run in real-time setting and also achieves a significant margin compared with previous real-time methods\cite{kirillov2019panoptic,sfnet,farSeg_iSAID}.

 \noindent \textbf{Visual Results Comparison:} 
 In Fig.~\ref{fig:vis_res_on_iSAID}, we compare our method results with several state-of-the-art methods~\cite{deeplabv3p,kirillov2019pointrend,kirillov2019panoptic} on the iSAID validation set. Our PFNet has better segmentation results on handling false positives of small objects and has fine-grain object mask boundaries.
 
\begin{table}[!t]\setlength{\tabcolsep}{6pt}
	\centering
	\begin{threeparttable}
		\scalebox{0.60}{
			\begin{tabular}{l | c c  |  c c}
				\toprule[0.15em]
			Method	 &  mIoU & mean-$F_1$ & mIoU & mean-$F_1$  \\
				\toprule[0.15em]
		PSPNet~\cite{pspnet} & 65.1 & 76.8 & 73.9 & 83.9\\
		FCN~\cite{fcn} & 64.2 & 75.9 & 73.1 & 83.1 \\
        OCnet(ASP-OC)~\cite{ocnet} & 65.7 & 77.4 &74.2 &84.1 \\
        Deeplabv3+~\cite{deeplabv3p} &  64.3 & 76.0  & 74.1 & 83.9 \\
        DAnet~\cite{DAnet} & 65.3 & 77.1 & 74.0 & 83.9\\
        CCnet\cite{ccnet} & 64.3 & 75.9 & 73.8 & 83.8 \\
        SemanticFPN~\cite{kirillov2019panoptic} & 66.3 & 77.6 & 74.3 & 84.0\\
        UPerNet~\cite{upernet} & 66.9 & 78.7 & 74.3 & 84.0 \\
        PointRend~\cite{kirillov2019pointrend} & 65.9 & 78.1 & 72.0 & 82.7 \\
        HRNet-W18~\cite{HRNet} & 66.9 & 78.2 & 73.4 & 83.4 \\
        GSCNN~\cite{gated-scnn} & 67.7 & 79.5 & 73.4 & 84.1 \\
        SFNet~\cite{sfnet} & 67.6 & 78.6 & 74.3 & 84.0\\
        EMANet~\cite{EMAnet} & 65.6 & 77.7 &72.9 & 83.1\\
        RANet~\cite{relation_aerial} & 66.1 &78.2 & 73.8 & 83.9 \\
        EncodingNet~\cite{context_encoding} & 65.5 & 77.4 & 73.4 & 83.5\\
        Denseaspp~\cite{denseaspp} & 64.7 & 76.4 & 73.9 & 83.9\\
        \hline
        PFNet  & \bf{70.4} & \bf{81.9}  & \bf{75.4} & \bf{84.8}  \\
	\bottomrule[0.1em]
	\end{tabular}}
		\caption{\small Comparison with the state-of-the-art results on Vahihigen(left) and Potsdam(right) datasets. 
		}
		\label{tab:results_on_aerial_datasets}
	\end{threeparttable}
\end{table}

\subsection{Results on general segmentation benchmarks:}

\noindent
We further verify our approach on general segmentation benchmarks including Cityscapes~\cite{Cityscapes}, ADE-20k~\cite{ADE20K} and BDD~\cite{yu2020bdd100k} for only verification purpose. We only report the results due to the limited space. More implementation details and visual results can be found in the supplementary file. We train both our baseline model and PFnet model on train datasets and report results on validation datasets under the same setting. 

\begin{table}[!t]\setlength{\tabcolsep}{6pt}
	\centering
	\begin{threeparttable}
		\scalebox{0.65}{
			\begin{tabular}{l c c c c c}
				\toprule[0.15em]
			Method	 &  Cityscapes & ADE20k & BDD & Param(M) & GFlops(G) \\
				\toprule[0.15em]
		PSPNet~\cite{pspnet} &  78.0 & 41.3 & 61.3 & 31.1 & 120.4 \\
        OCNet~\cite{ocnet} &  79.2 & 41.8 & 62.1 & 64.7 & 290.4 \\
        Deeplabv3+~\cite{deeplabv3p} &  79.4 & 42.0 & 61.0 &  40.5 & 189.8 \\
        \hline
		baseline +PPM &  78.8 & 40.9 & 61.1 & 32.9 & 83.1 \\
        Our PFnet  & 80.3 & 42.4 & 62.7 & 33.0 & 85.8  \\
	\bottomrule[0.1em]
	\end{tabular}}
		\caption{\small Experiment results on general datasets including Cityscapes, ADE20k, BDD validation datasets. All the methods are trained under the same training setting and the results are reported with single scale inputs. The GFlops is calculated with $512 \times 512 $ as input. All the methods use the ResNet50 backbone.}
		\label{tab:results_on_general_datasets}
	\end{threeparttable}
\end{table}

\noindent \textbf{Comparison with the Baseline Methods:} As shown in the last two rows of Tab.~\ref{tab:results_on_general_datasets}, our method improves the baseline model on various datasets about 1\% mIoU with fewer parameters and GFlops increase. Compared with the previous work~\cite{pspnet,ocnet,deeplabv3p}, our method achieves better results with much less computation cost.

%% file: 5conclusion.tex
\section{Conclusion}
In this paper, we propose PointFlow Module to solve both imbalanced foreground-background objects and semantic gaps between feature pyramids problems for aerial image segmentation. We design a novel Dual Point Matcher to sampled the matched points from salient areas and boundaries accordingly.
Extensive experiments have shown that our PF module can improve various baselines significantly on aerial benchmark. Based on the FPN framework, we build an efficient PFNet which achieves the best speed and accuracy trade-off on three public aerial benchmarks. Further experiments on three general segmentation datasets also prove the generality of our method.

%% file: 7ack.tex
\section{Acknowledgement}
Much thanks to the SenseTime Research's GPU servers for benchmarking semantic segmentation results on these aerial datasets. Y. Tong is supported by the National Key Research and Development Program of China (No.2020YFB2103402). Z. Lin is supported by NSF China (grant no.s 61625301 and 61731018), Major Scientific Research Project of Zhejiang Lab (grant no.s 2019KB0AC01 and 2019KB0AB02), Beijing Academy of Artificial Intelligence, and Qualcomm. 

%% file: 6sub.tex

\section{Supplementary}

\noindent \textbf{Overview:} In this supplementary, we will present more experiments details on Aerial datasets in the first section. Then we will provide detailed descriptions and visualization results on general semantic segmentation datasets.

\subsection{Experiments on Aerial Datasets}
In this section, we give supplementary aerial dataset experiments for the main paper due to the space limitation.

\noindent \textbf{Application on Various Methods:} For Deeplabv3~\cite{deeplabv3}, we append our PFNet decoder after the ASPP output which makes the total network as an encoder-decorder framework like U-net~\cite{unet}. For CCnet~\cite{ccnet}, we follow the same pipeline of Deeplabv3 by replacing ASPP head with CC-head. The GFlops in the table are calculated with $512 \times 512 $ image inputs.

\begin{figure*}[!h]
	\centering
	\includegraphics[width=0.80\linewidth]{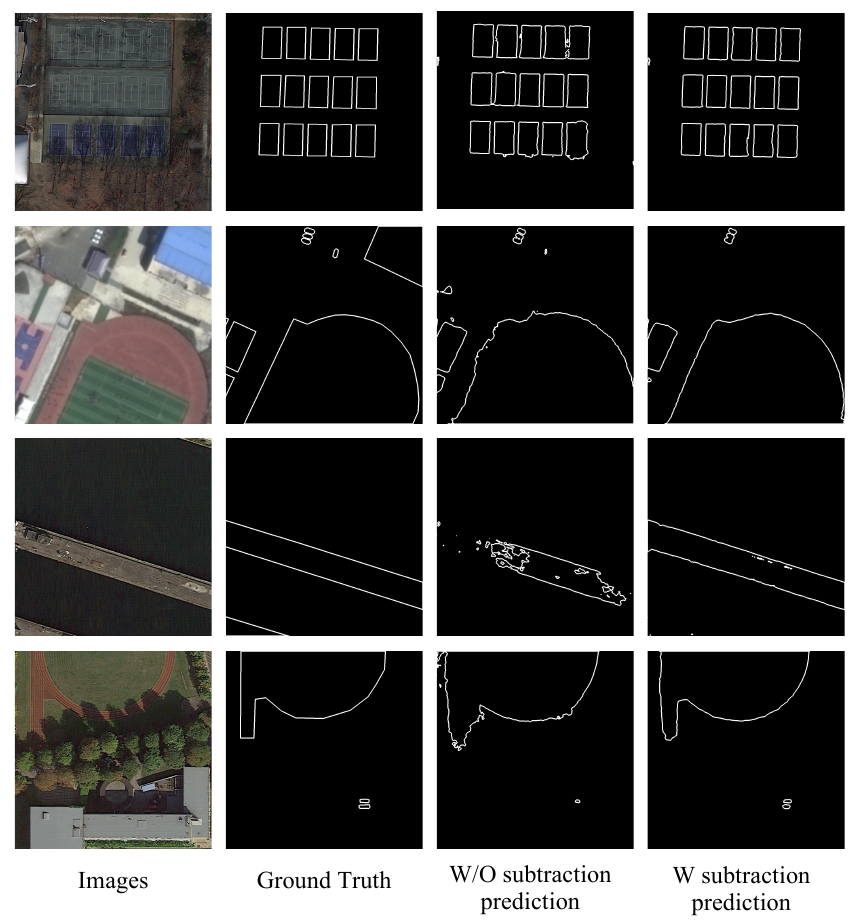}
	\caption{\small Visualization boundary results on iSAID validation dataset.  Best view it on screen.}
	\label{fig:iSAID_boundary_vis_sup}
\end{figure*}

\noindent \textbf{Effectiveness on Segmentation Boundaries:} We present the our subtraction based boundary prediction in Fig.~\ref{fig:iSAID_boundary_vis_sup}.
As shown in the figure, our subtraction based prediction leads to thinner and sharpen prediction in the first row and second row of Fig.~\ref{fig:iSAID_boundary_vis_sup} and avoids the inner part noise shown in the third row of Fig.~\ref{fig:iSAID_boundary_vis_sup}.

\noindent \textbf{Effect of Context Head:} We also verify the effectiveness of context head by replacing PPM~\cite{pspnet} into ASPP~\cite{deeplabv3} where we obtain 66.0 mIoU (0.9 mIoU drop compared with PPM head). Due to the efficiency of PPM, we choose it as our final context head.

\noindent \textbf{Foreground Points Ratio of Propagation:} We count the sampled points from the three different PFMs by finding the \textbf{intersection} of all matched points from Dual Point Matcher in each PFM. We match all the indexes into the high resolution map during the calculation. Thus the propagated points with no-overlap are considered.

\begin{figure*}[!h]
	\centering
	\includegraphics[width=1.0\linewidth]{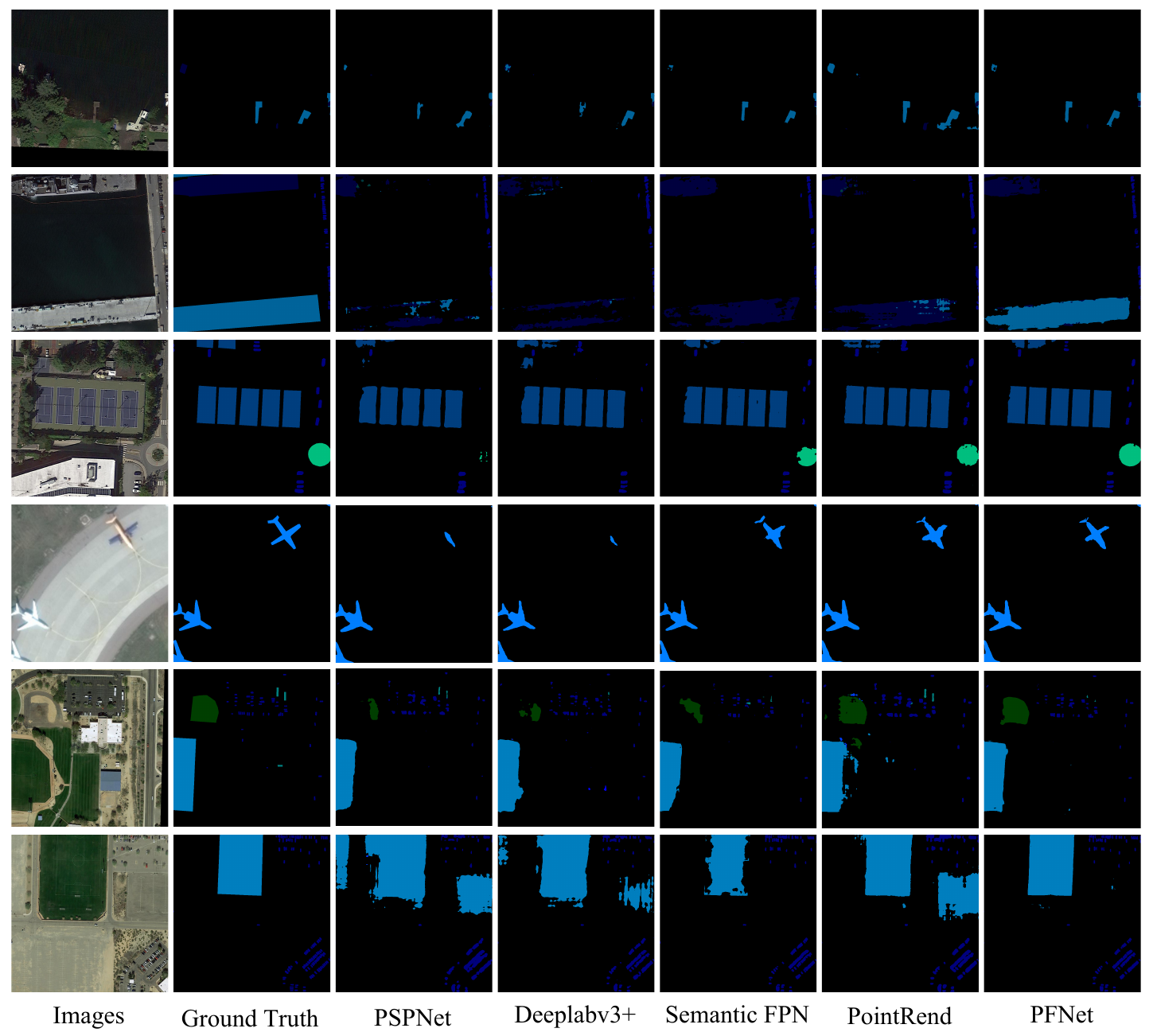}
	\caption{\small Visualization results on iSAID validation dataset.  Best view it on screen.}
	\label{fig:iSAID_vis_sup}
\end{figure*}

\begin{figure*}[!h]
	\centering
	\includegraphics[width=0.80\linewidth]{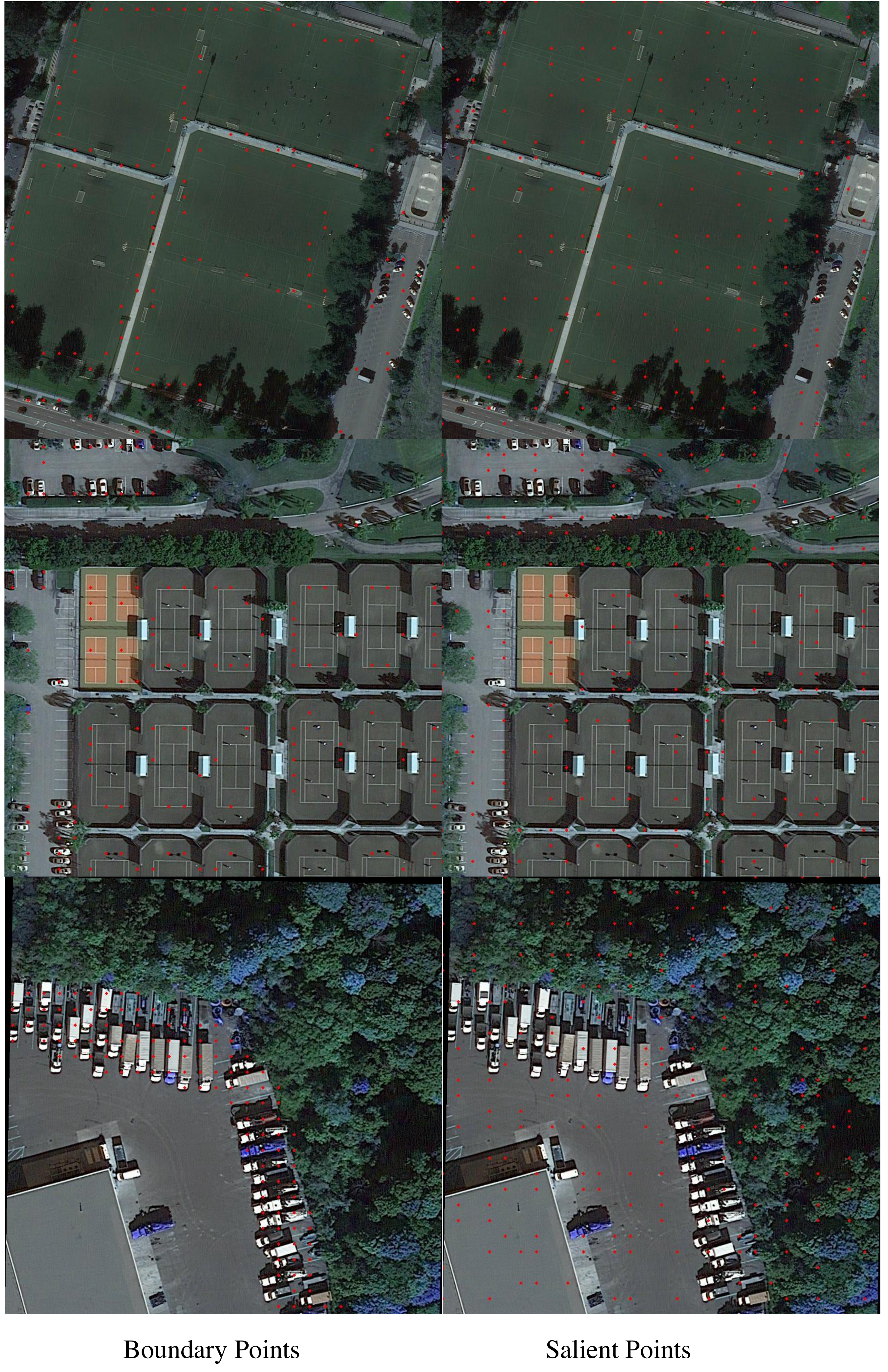}
	\caption{\small Visualization sampled points on iSAID validation dataset.  Best view it on screen.}
	\label{fig:iSAID_points_vis_sup}
\end{figure*}

\noindent \textbf{More Visualization on iSAID dataset:}
Fig.\ref{fig:iSAID_vis_sup} provides more visualization results on iSAID datasets.
Compared with previous work, our method has better segmentation results and less false positives in the scene. Fig.~\ref{fig:iSAID_points_vis_sup} shows more matched points visualization on full cropped images.

\noindent \textbf{Detailed Results:} 
Tab.~\ref{tab:results_iSAID_dataset} reports the per-class results on iSAID datasets. From that table, our PFNet achieves best results on 14 classes (total 15 classes). Tab.~\ref{tab:results_Postdam_dataset} and Tab.~\ref{tab:results_Vaihingen_dataset} report per-class results on Postdam and Vaihingen dataset receptively. Our PFNet also achieves top performance.

\noindent \textbf{Dataset Split on Potsdam and Vaihingen:}
We provide the detailed dataset split as following:

\noindent
The Potsdam dataset consists of 38 high resolution aerial images. To train and evaluate networks, we utilize 24 images for training and 14 images for testing. For training set, the image IDs are 2\_10, 2\_11, 2\_12, 3\_10, 3\_11, 3\_12, 4\_10, 4\_11, 4\_12, 5\_10, 5\_11, 5\_12, 6\_7, 6\_8, 6\_9, 6\_10, 6\_11, 6\_12, 7\_7, 7\_8, 7\_9, 7\_10, 7\_11, 7\_12. For test set, the image IDs are  2\_13, 2\_14, 3\_13, 3\_14, 4\_13, 4\_14, 4\_15, 5\_13, 5\_14, 5\_15, 6\_13, 6\_14, 6\_15, 7\_13.

\noindent
The Vaihingen dataset has 33 high resolution aerial images. We selects 16 images for trainand and another 17 images as the testset to evaluate the models. The image IDs for training set are 1, 3, 5, 7, 11, 13, 15, 17, 21, 23, 26, 28, 30, 32, 34, 37 and image IDs for test set are 2, 4, 6, 8, 10, 12, 14, 16, 20, 22, 24, 27, 29, 31, 33, 35, 38.

\noindent \textbf{Speed test details:} We test our models on single V100 GPU by averaging inference time with 100 image. The Pytorch version is 1.5 with CUDA-10.1.

\begin{table*}\setlength{\tabcolsep}{4pt}
	\centering
	\begin{threeparttable}
		\scalebox{0.75}{
			\begin{tabular}{l | c | c | c c c c c c c c c c c c c c c}
				\toprule[0.15em]
				\multirow{2}{*}{Method} & \multirow{2}{*}{backbone} & \multirow{2}{*}{mIoU(\%)} & \multicolumn{14}{c}{IoU per category(\%)} &   \\
				\cline{4-18}
				\multirow{1}{*}{ } & \multirow{1}{*}{ } & \multirow{1}{*}{ } & Ship & ST & BD &TC &BC &GTF &Bridge & LV &SV &HC &SP & RA &SBF &Plane &Harbor \\
				\hline
				DenseASPP~\cite{denseaspp} & RenNet50 & 57.3 & 55.7 & 63.5 & 67.2 & 81.7 & 54.8 & 52.6 & 34.7 & 55.6 & 36.3 & 33.4 & 37.5 & 53.4 & 73.3 & 74.7 & 46.7  \\
                RefineNet ~\cite{refinenet} & ResNet50 & 60.2  & 63.8 & 58.6 & 72.3 & 85.3 & 61.1 & 52.8 & 32.6 & 58.2 & 42.4 & 23.0 & 43.4 & 65.6 & 74.4 & 79.9 & 51.1 \\
        		PSPNet~\cite{pspnet} & ResNet50 & 60.3 & 65.2 & 52.1 & 75.7 & 85.6 & 61.1 & 60.2 & 32.5 & 58.0 & 43.0 & 10.9 & 46.8 & 68.6 & 71.9 & 79.5 & 54.3\\
                OCNet-(ASP-OC)~\cite{ocnet} & ResNet50 &  40.2 & 47.3 & 40.2 & 44.4 & 65.0 & 24.1 & 29.9 & 2.71 & 46.3 & 13.6 & 10.3 & 34.6 & 37.9 & 41.4 & 68.1 & 38.0 \\
                EMANet~\cite{EMAnet} & ResNet50 & 55.4 & 63.1 & 68.4 & 66.2 & 82.7 & 56.0 & 18.8 & 42.1 & 58.2 & 41.0 & 33.4 & 38.9 & 46.9 & 46.4 & 78.5  & 47.5 \\
                CCNet~\cite{ccnet} & ResNet50 & 58.3 & 61.4 & 65.7 & 68.9 & 82.9 & 57.1 & 56.8 & 34.0 & 57.6 & 38.3 & 31.6 & 36.5 & 57.2 & 75.0 & 75.8 & 45.9\\
                EncodingNet~\cite{context_encoding} & ResNet50 & 58.9 & 59.7 & 64.9 & 70.0 & 84.2 & 55.2 & 46.3 & 36.8 & 57.2 & 38.7 & 34.8 & 42.4 & 59.8& 69.8 & 76.1 & 48.0  \\
                SemanticFPN~\cite{kirillov2019panoptic} & ResNet50 & 62.1 & 68.9 & 62.0 & 72.1 & 85.4 & 54.1 & 48.9 & 44.9 & 61.0 & 48.6 & 37.4 & 42.8 & 70.2 & 58.6 & 84.7 & 54.9 \\
                UPerNet~\cite{kirillov2019panoptic} & ResNet50 & 63.8 & 68.7 & 71.0 & 73.1 & 85.5 & 55.3 & 57.3 & 43.0 & 61.3 & 45.6 & 30.3 & 45.7 & 68.7 & 75.1 & 84.3 & 56.2 \\
                HRNet\cite{upernet} & HRNetW18 & 61.5 & 65.9 & 68.9 & 74.0 & 86.9 & 59.4 & 61.5 & 33.8 & 62.1 & 46.9 & 14.9 & 44.2 & 52.9 & 75.6 & 81.7 & 52.2 \\
                SFNet\cite{sfnet} & ResNet50 & 64.3 & 68.8 & 71.3 & 72.1 & 85.6& 58.8 & 60.9 & 43.1 & 62.9 & 47.7 & 30.4 & 47.8 & 69.8 & 75.1 & 83.1 & 57.3 \\
                GSCNN\cite{gated-scnn} & ResNe50 & 63.4 & 65.9 & 71.2 & 72.6 & 85.5 & 56.1 & 58.4 & 40.7 & 63.8 & 51.1 & 33.8 & 48.8 & 58.5 & 72.5 & 83.6 & 54.4 \\
                RANet\cite{relation_aerial} & ResNet50 & 62.1 & 67.1 & 61.3 & 72.5 & 85.1 & 53.2 & 47.1 & 45.3 & 60.1 & 49.3 & 38.1 & 41.8 & 70.5 & 58.8 & 83.1 & 55.6  \\
                FarSeg~\cite{farSeg_iSAID} & ResNet50 & 63.7 & 65.4 & 61.8 & 77.7 & 86.4 & 62.1 & 56.7 & 36.7 & 60.6 & 46.3 & 35.8 & \bf{51.2} & 71.4 & 72.5 & 82.0 & 53.9 \\
                \hline
                PFNet  & ResNet50 & \bf{66.9} & \bf{70.3} & \bf{74.7} & \bf{77.8} &  \bf{87.7} & \bf{62.2} &  \bf{59.5} & \bf{45.2} & \bf{64.6} & \bf{50.2} & \bf{37.9} & 50.1 & \bf{71.7} & \bf{75.4} & \bf{85.0} & 
                \bf{59.3} \\
        
	\bottomrule[0.1em]
	\end{tabular}}
		\caption{Experimental results on iSAID $val$ set. The bold values in each column mean the best entries. The category are defined as: ship (Ship), storage tank (ST), baseball court (BC), ground field track (GTF), bridge (Bridge), large vehicle (LV), small vehicle (SV), helicopter (HC), swimming pool (SP), roundabout (RA), soccerball field (SBF), plane (Plane), harbor (Harbor). All the models are trained under the same setting following the FarSeg~\cite{farSeg_iSAID}.}
		\label{tab:results_iSAID_dataset}
	\end{threeparttable}
\end{table*}

\begin{table*}\setlength{\tabcolsep}{6pt}
	\centering
	\begin{threeparttable}
		\scalebox{0.9}{
			\begin{tabular}{l | c | c | c c c c c c}
				\toprule[0.15em]
				\multirow{2}{*}{Method} & \multirow{2}{*}{mIoU(\%)} & \multirow{2}{*}{mean-$F_1$} & \multicolumn{5}{c}{$F_1$ per category} &   \\
				\cline{4-9}
				\multirow{1}{*}{ } & \multirow{1}{*}{ } & \multirow{1}{*}{ } & Imp.surf. & Build. & Low veg. &Tree & Car & Cluster \\
				\hline
		PSPNet~\cite{pspnet}  &   65.1 & 76.8 & 88.4 & 92.8 & 79.2 & 85.9 & 73.5 & 41.0 \\
    	FCN~\cite{fcn}  &  64.2 & 75.9 & 87.6  & 91.6 & 77.8 & 84.6 & 73.5 & 40.3 \\
        OCNet(ASP-OC)~\cite{ocnet}   &  65.7 & 77.4 & 88.8 & 92.9 & 79.2 & 85.8 & 73.9 & 43.8 \\
        Deeplabv3+~\cite{deeplabv3p} & 64.3 & 76.0 & 88.7 & 92.8 & 78.9  & 85.6 & 72.4 & 37.6 \\
        DANet~\cite{DAnet} & 65.3 & 77.1 & 88.5 & 92.7 & 78.8 & 85.7 & 73.7 & 43.2 \\
        CCNet~\cite{ccnet} & 64.3 & 75.9 & 88.3 & 92.5 & 78.8 & 85.7 & 73.9 & 36.3 \\
        SemanticFPN~\cite{kirillov2019panoptic} & 66.3 & 77.6 & 89.6 & 93.6 & 79.7 & \bf{86.3} & 75.7 & 40.7\\
        UPerNet~\cite{upernet} & 66.9 & 78.7 & 89.2 & 93.0 & 79.4 & 86.0 & 74.9 & 49.7 \\
        PointRend~\cite{kirillov2019pointrend} & 65.9 & 78.1 & 88.2 & 92.4 & 78.9 & 84.5 & 73.5 & 51.1 \\
        HRNet-W18~\cite{HRNet} & 66.9 & 78.2 & 89.2 & 92.6 & 78.7 & 85.7 & 77.1 & 45.9 \\
        GSCNN~\cite{gated-scnn} & 67.7 & 79.5 & 89.4 & 92.6 & 78.8 & 85.4 & 77.9 & 52.9 \\
        SFNet~\cite{sfnet} & 67.6 & 78.6 & 90.0 & \bf{94.0} & \bf{80.3} & 86.5 & 78.9 & 41.9\\
        EMANet~\cite{EMAnet} & 65.6 & 77.7 & 88.2 & 92.7 & 78.0 & 85.7 & 72.7 & 48.9\\
        RANet~\cite{relation_aerial} & 66.1 & 78.2 & 88.0 & 92.3 & 79.1 & 86.0 & 78.8 & 53.1  \\ 
        EncodingNet~\cite{context_encoding} & 65.5 & 77.4 & 88.6 & 92.5 & 78.5 & 85.7 & 73.6 & 45.5\\
        Denseaspp~\cite{denseaspp} & 64.7 & 76.4 & 87.3 & 91.1 & 76.2 & 83.4 & 77.1 & 43.3 \\
        \hline
        PFNet & \bf{70.4} & \bf{81.9} & \bf{90.1} & 93.6 & 77.7 & 85.4 & \bf{80.0} & \bf{64.6} \\
        
	\bottomrule[0.1em]
	\end{tabular}}
		\caption{Experimental results on the Vaihingen Dataset. The results are reported with single scale input. The bold values in each column mean the best entries. The category are defined as: impervious surfaces (Imp.surf.), buildings (Build), low vegetation (Low veg), trees (Tree), cars (Car), cluster/background (Cluster).}
		\label{tab:results_Vaihingen_dataset}
	\end{threeparttable}
\end{table*}

\begin{table*}\setlength{\tabcolsep}{6pt}
	\centering
	\begin{threeparttable}
		\scalebox{0.9}{
			\begin{tabular}{l | c | c | c c c c c c}
				\toprule[0.15em]
				\multirow{2}{*}{Method} & \multirow{2}{*}{mIoU(\%)} & \multirow{2}{*}{mean-$F_1$} & \multicolumn{5}{c}{$F_1$ per category} &   \\
				\cline{4-9}
				\multirow{1}{*}{ } & \multirow{1}{*}{ } & \multirow{1}{*}{ } & Imp.surf. & Build. & Low veg. &Tree & Car & Cluster \\
				\hline
	PSPNet~\cite{pspnet} & 73.9 & 83.9 & 90.8 & 95.4 & 84.5 & 86.1 & 88.6 & 58.0 \\
		FCN~\cite{fcn} & 73.1 & 83.1  & 90.2 & 94.7 & 84.1 & 85.6 & 89.2 & 54.8 \\
        OCnet(ASP-OC)~\cite{ocnet} & 74.2 & 84.1 & 90.9 & 95.5 & 84.8 & 86.0 & 89.2 & 58.2 \\
        Deeplabv3+~\cite{deeplabv3p} & 74.1 & 83.9 & 91.0 & 95.6 & 84.6 & 86.0 & 90.0 & 56.2 \\
        DAnet~\cite{DAnet} & 74.0 & 83.9 & 91.0 & 95.6 & 84.9 & 86.2 & 88.7 & 57.0 \\
        CCnet\cite{ccnet} & 73.8 & 83.8 &  90.7 & 95.5 & 84.7 & 86.0 & 88.5 & 57.3 \\
        SemanticFPN~\cite{kirillov2019panoptic} & 74.3 & 84.0 & 91.0 & 95.5 & 84.9 & 85.9 & 90.4 & 56.3 \\
        UPerNet~\cite{upernet} & 74.3 & 84.0 & 90.9 & 95.7 & 85.0 & 86.0 & 90.2 & 56.2 \\
        PointRend~\cite{kirillov2019pointrend} & 72.0 & 82.7 & 89.8 & 94.6 & 82.8 & 85.2 & 85.2 & 58.6 \\
        HRNet-W18~\cite{HRNet} & 73.4 & 83.4 & 90.4 & 94.9 & 84.2 & 85.4 & 90.0 & 55.5 \\
        GSCNN~\cite{gated-scnn} & 73.4 & 84.1 & 91.4 & 95.5 & 84.8 & 85.8 & \bf{91.2} & 55.9 \\
        SFNet~\cite{sfnet} & 74.3 & 84.0 & 91.0 & 95.5 & 85.1 & 86.0 & 90.9 & 55.5 \\
        EMANet~\cite{EMAnet} & 72.9 & 83.1 & 90.4 & 94.9 & 84.2 & 85.7 & 88.3 & 55.1 \\
        RANet~\cite{relation_aerial} & 73.8 & 83.9 & 90.8 & 92.1 & 84.3 & 86.8 & 88.9 & 56.0 \\
        EncodingNet~\cite{context_encoding} & 73.4 & 83.5 & 90.6 & 95.1 & 84.5 & 86.0 & 88.2 & 56.6 \\
        Denseaspp~\cite{denseaspp} & 73.9 & 83.9 & 90.8 & 95.4 & 84.6 & 86.0 & 88.5 & 58.1 \\
        \hline
        PFNet  & \bf{75.4} & \bf{84.8} & \bf{91.5} & \bf{95.9} & \bf{85.4} & \bf{86.3} & 91.1 & \bf{58.6} \\
        
	\bottomrule[0.1em]
	\end{tabular}}
		\caption{Experimental results on Postdam Dataset. The results are reported with single scale input. The bold values in each column mean the best entries. The category are defined as: impervious surfaces (Imp.surf.), buildings (Build), low vegetation (Low veg), trees (Tree), cars (Car), cluster/background (Cluster).}
		\label{tab:results_Postdam_dataset}
	\end{threeparttable}
\end{table*}

\subsection{Experiments on Generation Segmentation Dataset}
In this section, we will first give the implementation details on Cityscapes~\cite{Cityscapes}, BDD~\cite{yu2020bdd100k} and ADE20k~\cite{ADE20K}.

\noindent \textbf{Implementation Details}

\noindent
\textbf{Cityscapes} dataset is composed of a large set of high-resolution $(2048 \times 1024)$ images in street scenes. This dataset has 5,000 images with high quality pixel-wise annotations for 19 classes, which is further divided into 2975, 500, and 1525 images for training, validation and testing. We only use the fine-data for training. Following the common practice, the ``poly'' learning rate policy is adopted to decay the initial learning rate by multiplying $(1 -\frac{\text{iter}}{\text{total}\_\text{iter}})^{0.9}$ during training. Data augmentation contains random horizontal flip, random resizing with scale range of $[0.75, ~2.0]$, and random cropping with crop size of $864 \times 864$ and we train totally 300 epochs for strong baseline.

\noindent
\textbf{BDD} is a new road scene benchmark consisting $7,000$ images for training and $1,000$ images for validation. We follow the same setting as Cityscapes dataset.

\noindent
\textbf{ADE20k} is a more challenging scene parsing dataset annotated with 150 classes, and it contains 20K/2K images for training and validation. It has the various objects in the scene. We train the network for 120 epochs with batch size 16, crop size 512 and initial learning rate 1e-2. For final testing, we perform multi-scale testing with horizontal flip operation.

\begin{figure*}[!h]
	\centering
	\includegraphics[width=1.0\linewidth]{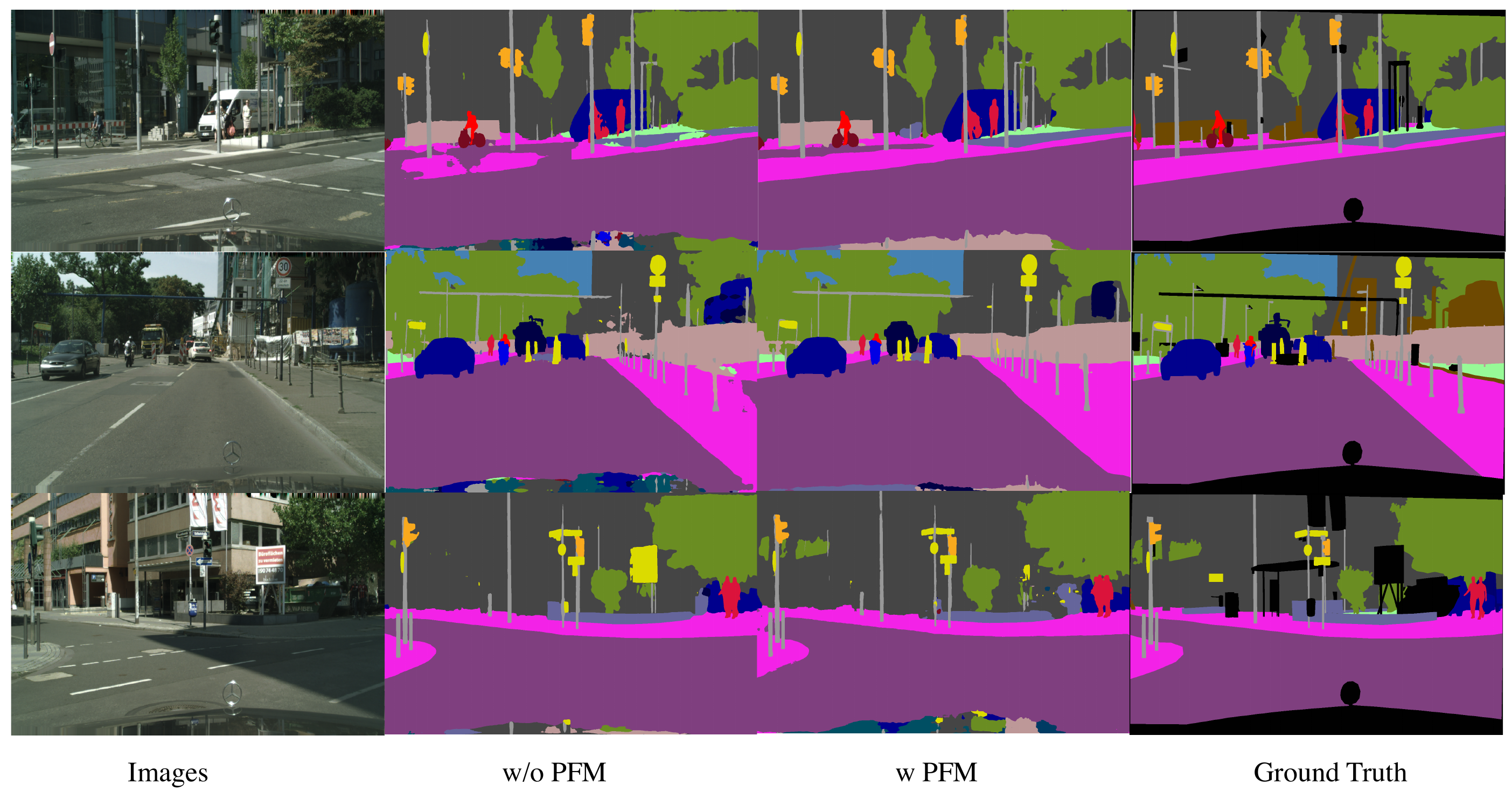}
	\caption{Visualization results on Cityscapes dataset.  Best view it on screen.
	}
	\label{fig:city_vis_sup}
\end{figure*}

\begin{figure*}[hp]
	\centering
	\includegraphics[width=0.8\linewidth]{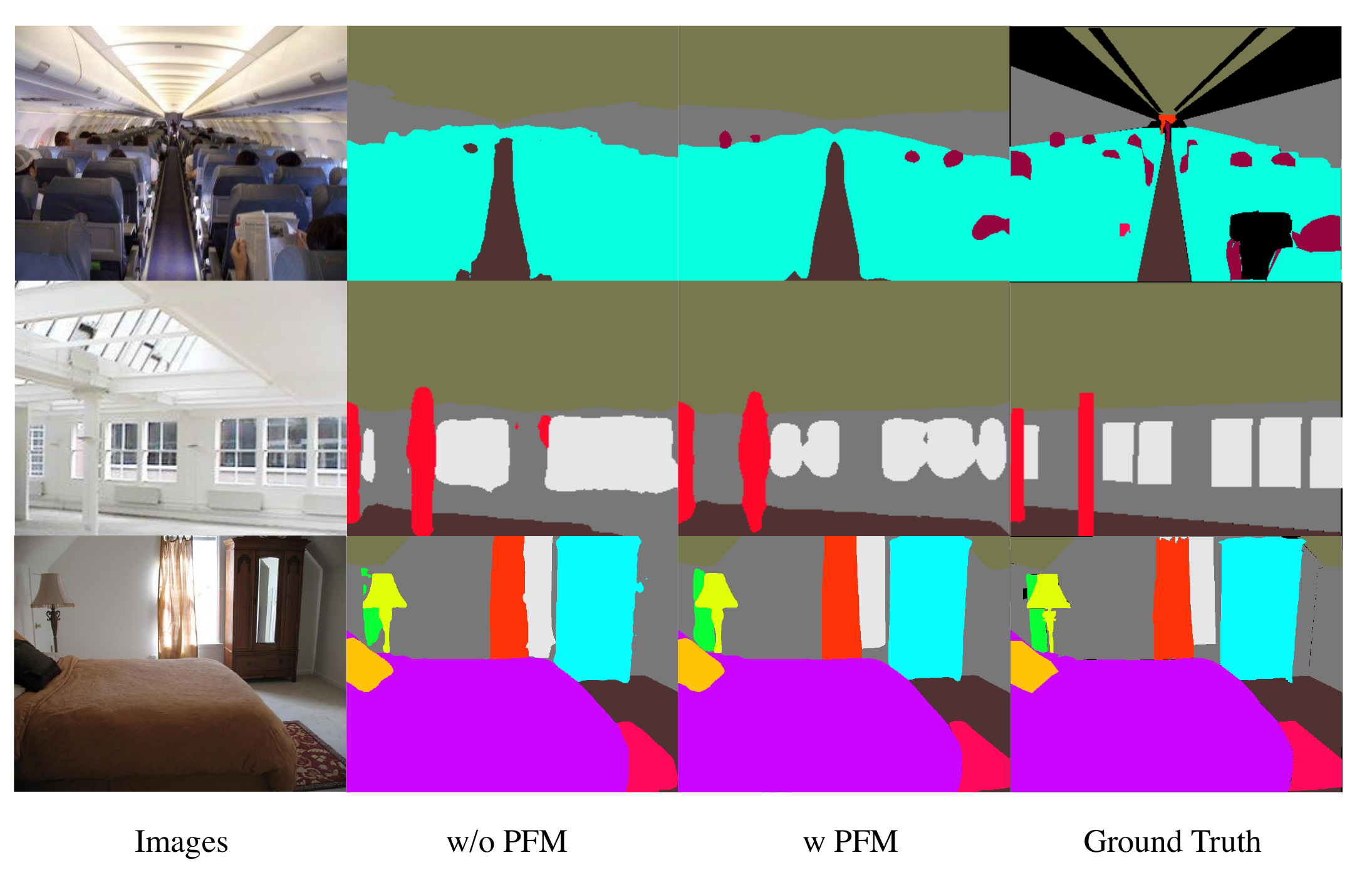}
	\caption{Visualization results on ADE20k dataset. Best view it on screen.
	}
	\label{fig:ade_vis_sup}
\end{figure*}

\noindent \textbf{Visualization and Comparison Results}
We provide visualization and comparison results on Cityscapes and ADE-20k in Fig.~\ref{fig:city_vis_sup} and Fig.~\ref{fig:ade_vis_sup}. From both figures, our PFM can better handle the segmentation tails and missing objects in the scene.

